\RequirePackage{fix-cm}
\documentclass[twocolumn]{svjour3}          
\smartqed  
\usepackage{amsmath}
\usepackage{graphicx,array}
\usepackage{setspace}
\usepackage{xcolor}
\usepackage{color,soul}
\usepackage{relsize}
\usepackage{multirow}
\usepackage{tabularx}
\usepackage{subfigure}
\usepackage{booktabs}
\usepackage[hidelinks]{hyperref}
\usepackage{flushend}
\graphicspath{{fig/}}

\journalname{International Journal on Digital Libraries}

\newcommand{\rowindent}{\hspace{2mm}}

\DeclareMathOperator{\issyn}{is-syn}

\begin{document}

\title{Scientific document summarization via citation contextualization and scientific discourse \thanks{
$^*$ This is a pre-print of an article published on IJDL. The final publication is available at Springer via http://dx.doi.org/10.1007/s00799-017-0216-8}
}



\author{Arman Cohan $^1$        \and
        Nazli Goharian  $^1$
}


\institute{ Arman Cohan
\at
              \email{arman@ir.cs.georgetown.edu}           
           \and
           Nazli Goharian
            \at
              \email{nazli@ir.cs.georgetown.edu}
          \and
          $^1$ Information Retrieval Lab, Department of Computer Science, Georgetown University, Washington DC, USA
}

\date{Received: date / Accepted: date}
%




\maketitle

 \sloppy
\begin{abstract}
The rapid growth of scientific literature has made it difficult for the researchers to quickly learn about the developments in their respective fields. Scientific document summarization addresses this challenge by providing summaries of the important contributions of scientific papers. We present a framework for scientific summarization which takes advantage of the citations and the scientific discourse structure. Citation texts often lack the evidence and context to support the content of the cited paper and are even sometimes inaccurate. We first address the problem of inaccuracy of the citation texts by finding the relevant context from the cited paper. We propose three approaches for contextualizing citations which are based on query reformulation, word embeddings, and supervised learning. We then train a model to identify the discourse facets for each citation. We finally propose a method for summarizing scientific papers by leveraging the faceted citations and their corresponding contexts. We evaluate our proposed method on two scientific summarization datasets in the biomedical and computational linguistics domains. Extensive evaluation results show that our methods can improve over the state of the art by large margins.
\end{abstract}


\section{Introduction}
\label{sec:intro}

The rapid growth of scientific literature in recent decades has created a challenge for researchers in various fields to keep up with the newest developments. According to a recent study by bibliometric analysts, the global scientific output doubles approximately every nine years \cite{bornmann2015growth}, further signifying this challenge. Existence of surveys in different fields shows that finding an overview of key developments in scientific areas is desirable, however procuring such surveys requires extensive human efforts. Scientific summarization aims at addressing this problem by providing a concise representation of important findings and contributions of scientific papers, reducing the time required to overview the entire paper to understand important contributions. Article abstracts are a basic form of scientific summaries. While abstracts provide an overview of the paper, they do not necessarily convey all the important contributions and impacts of the paper \cite{elkiss2008blind}: (\textit{i}) The authors might ascribe contributions to their papers that are not existent. (\textit{ii}) some important contributions might not be included in the abstract. (\textit{iii}) the contributions stated in the abstract do not convey the article's impact over time. (\textit{iv}) abstracts usually provide a very broad view of the papers and they may not be detailed enough for people seeking detailed contributions. (\textit{v}) The content distribution in the abstracts are not evenly drawn from different sections of the papers \cite{atanassova2016composition}. These problems have inspired another type of scientific summaries which are obtained by utilizing a set of citations referencing the original paper \cite{qazvinian2008scientific,qazvinian2013generating}. Each citation, is often accompanied by a short description, explaining the ideas, methods, results, or findings of the cited work. This short description is called citation text or citance \cite{nakov2004citances}. Therefore, a set of citation texts by different papers can provide an overview of the main ideas, methods and contributions of the cited paper, and thus, can form a summary of the referenced paper. These community based summaries capture the important contributions of the paper, view the article from multiple aspects, and reflect the impact of the article to the community.

At the same time, there are multiple problems associated with citation texts. They are written by different authors so they may be biased toward another work. The citation texts lack the context in terms of the details of the methods, the data, assumptions, and results. More importantly, the points and claims by the original paper might be misunderstood by the citing authors; certain contributions might be ascribed to the cited work that are not on par with the original author's intent. Another serious problem is the modification of the epistemic value of claims, which states that many claims by the original author might be stated as facts in the future citations \cite{deWaard2012epistemic}. An example of this is shown in Figure \ref{fig:epistemic_drift}. As illustrated, while the original authors write on some possibilities, later the citing authors state them as known facts. These problems are even more serious in biomedical domain where slight misrepresentations of the specific findings about treatments, diagnosis, and medications, could directly affect human lives.

\begin{figure}[t]
        \begin{minipage}{\columnwidth}
        \center
            {\small \textbf{Reference Article}}
                        \vspace{2pt}
        \end{minipage}
        \begin{tabular}{|p{.94\columnwidth}|}
            \hline
            (Voorhoeve et al., 2006): ``These miRNAs could neutralize p53-mediated CDK inhibition, \textcolor{red}{possibly} through direct inhibition of the expression of the tumor suppressor LATS2.'' \\
            \hline
        \end{tabular}
        \begin{minipage}{\columnwidth}
        \center
            \vspace{2pt}
            {\small \textbf{Citing Articles}}
            \vspace{2pt}
        \end{minipage}
        \begin{tabular}{|p{.94\columnwidth}|}
            \hline
            (Kloosterman and Plasterk, 2008): ``In a genetic screen, miR-372 and miR-373 \textcolor{red}{were found to} allow proliferation of primary human cells (Voorhoeve et al., 2006).'' \\ \hline
        \end{tabular}
        \begin{tabular}{|p{.94\columnwidth}|} \hline
            (Okada et al., 2011): ``Two oncogenic miRNAs, miR-372 and miR-373, \textcolor{red}{directly inhibit} the expression of Lats2, \textcolor{red}{thereby} allowing tumorigenic growth in the presence of p53 (Voorhoeve et al., 2006).''
            \\ \hline
        \end{tabular}

        {\footnotesize\caption{\label{fig:epistemic_drift} Example of epistemic value drift \cite{deWaard2012epistemic}. The claims that Voorhoeve et al. (2006) state as possibilities, becomes fact in later citations (Okada et al., 2011; Kloosterman and Plasterk, 2008).}}
\vspace{-12pt}
\end{figure}

One way to address such problems is to consider the citations in their context from the reference article. Therefore, citation texts should be linked to the specific parts in the reference paper that correctly reflect them. We call this ``citation contextualization''. Citation contextualization is a challenging task due to the terminology variations between the citing and cited author's language usage.

Scientific papers have the unique characteristic of following a specific discourse structure. For example, a typical scientific discourse structure follows this form: problem and motivation, methods, experiments, results, and implications. The rhetorical status of a citation provides additional useful information that can be used in applications such as information extraction, retrieval, and summarization \cite{Teufel2006}. Each citation text could refer to specific discourse facets of the referenced paper. For example one citation could be about the main method of the referenced paper while the other one could mention their results. Identifying these discourse facets has distinct values for scientific summarization; it allows creating more coherent summaries and diversifying the points included in the generated scientific summaries.

Scientific summarization is recently further motivated by TAC\footnote{Text Analysis Conference, http://tac.nist.gov/2014/BiomedSumm/} 2014 summarization track, and the 2016 computation linguistics summarization shared task \cite{jaidka2016overview}. Following these works and motivated by the challenges mentioned above, we propose a framework for scientific summarization based on citations. Our approach consists of the following steps:

\begin{itemize}
  \item \textit{Contextualizing citation texts}: We propose several approaches for contextualizing citations. Finding the exact reference context for the citations is challenging due to discourse variation and terminology differences between the citing and the referenced authors. Therefore, traditional Information Retrieval (IR) methods are inadequate for finding the relevant contexts. We propose to address this challenge by query reformulations, utilizing word embeddings \cite{bengio2003neural}, and domain-specific knowledge. Our main approach is a retrieval model for finding the appropriate context of the citations and is designed to handle terminology variations between the citing and cited authors.

  \item \textit{Discourse structure}: After extracting the context of the citation texts, we classify them into different discourse facets. We use a linear classifier with variety of features for classifying the citations.

  \item \textit{Summarization}: We propose two approaches for summarizing the papers. Both approaches are based on summarization through the scientific community where the main points of a paper are captured by a set of given citations. Our approach extends the previous works on citation-based summarization \cite{Qazvinian2013,qazvinian2008scientific,qazvinian2010identifying} by including the reference context to address the inaccuracy problem associated with the citation texts. After extracting the citation contexts from the reference paper, we group them into different discourse facets. Then using the most central sentences in each group, we generate the final summary.
\end{itemize}

In particular our contributions are summarized as follows: (\textit{i}) An approach for extracting the context of the citation texts from the reference article. (\textit{ii}) Identifying the discourse facets of the citation contexts. (\textit{iii}) A scientific summarization framework utilizing citation contexts and the scientific discourse structure. (\textit{iv}) Extensive evaluation on two scientific domains.


\section{Related work}
\label{sec:related}

\subsection{Citation text analysis}
\label{subsec:citation-text-analysis}

Citations play an integral role in the scientific development. They help disseminate the new findings and they allow new works to be grounded on previous efforts \cite{hernandez2016survey}. While there is a large body of related work on analysis of citation networks, instead of link analysis, we focus on textual aspects of the citations. To better utilize the citations, researchers have explored ways to extract citation texts, which are short textual parts describing some aspects of the cited work. Examples of the proposed approaches for extracting the citation texts include jointly modeling the link information and the citation texts \cite{kataria2010utilizing}, supervised Markov Random Fields classifiers \cite{qazvinian2010identifying}, and sequence labeling with segment classification \cite{abu2012reference}. These approaches focus on finding the sentences or textual spans in the citing article that explain some aspects of the
cited work. In this work, we assume that citation texts are already obtained either manually or by using one of these works. Given the citation texts, we instead focus on contextualizing these citation texts using the reference; we find the text spans in the reference article that most closely reflect the citation text.

There exists some related work on further analyzing the citations for finding their function or rhetorical status \cite{Teufel2006,garzone2000towards,abu2013purpose,hernandez2016survey}. In these works, the authors tried to identify the reasons behind citations which can be a statement of weakness, contrast or comparison, usage or compatibility, or a neutral category. They proposed a classification framework based on lexically and linguistically inspired features for classifying citation functions. The distribution of citations within the structure of scientific papers have been also studied \cite{Bertin2016}. The authors of \cite{chakraborty-narayanam:2016:EMNLP2016} have investigated the problem of measuring the intensity of the citations in scientific papers and in \cite{chakraborty2016ferosa}, the authors proposed using the discourse facets for scientific article recommendation. Recently, a framework for understanding citation function has been proposed \cite{jurgens2016citation} which unifies all the previous efforts in terms of definition of citation functions. While citation function can provide additional information for summarization, in this work we do not utilize these information. Instead, we utilize the discourse facet of the citation contexts in a reference paper.

\subsection{Citation contextualization}
\label{subsec:citation-context}

More recently, there has been some efforts in contextualizing citations from the reference. In particular, TAC 2014 summarization track,\footnote{http://tac.nist.gov/2014/BiomedSumm/} and the CL-SciSumm 2016 shared task on computational linguistic summarization \cite{jaidka2016overview} have released datasets to promote research for citation contextualization. The former is more domain specific, focusing on biomedical scientific literature, while the latter is in a more general domain consisting of publications in computational linguistics. To our knowledge, there is no overview paper on TAC. We briefly discuss the successful approaches in CL-SciSumm 2016. The authors of \cite{Cao2016PolyUAC} used an SVM-rank approach with features such as tf-idf\footnote{Term Frequency - Inverted Document Frequency.} cosine similarity, position of the reference sentence, section position, and named entity features. In another approach \cite{Li2016CISTSF}, the authors used an SVM classifier with sentence similarity and lexicon based features. The authors of \cite{Nomoto2016NEALAN} proposed a hybrid model based on tf-idf similarity and a single layer neural network that scores the relevant reference texts above the irrelevant ones. Finally, in the work by \cite{klampfl2016identifying}, the authors proposed the use of TextSentenceRank algorithm which is an enhanced version of the TextRank algorithm for ranking keywords in the documents. Here, we specifically focus on the problem of terminology variation between the citing and cited authors. We propose approaches that address this problem. Our proposed approaches are based on query reformulations, word embeddings, and domain-specific knowledge.

\subsection{Text summarization}
\label{subsec:summ}

Document summarization has been an active research area in NLP in recent decades; there is a rich literature on text summarization. Approaches towards summarization can be divided into the following categories: \textit{(i)} topic modeling based \cite{gong2001lsa,steinberger2004lsa,vanderwende2007beyond,celikyilmaz2010hybrid}: In these approaches, the content or topical distribution of the final summary is estimated using a probabilistic framework. (\textit{ii}) solving an optimization problem \cite{Clarke:2008,berg2011jointly,DurrettBergKlein2016}: these approaches cast the summarization problem as an optimization problem where an objective function needs to be optimized with respect to some constraints. (\textit{iii}) supervised models \cite{osborne2002using,conroy2011classy,Chali:2012:QMS:2139643.2139649}, where selection of sentences in the summary are learned using a supervised framework. (\textit{iv}) graph based \cite{erkan2004,mihalcea2004,Paul:2010}: these approaches seek to find the most central sentences in a document's graph where sentences are nodes and edges are similarities. (\textit{v}) Heuristic based \cite{carbonell1998use,guo2010probabilistic,lin2010putting}: these works approach the summarization problem by greedy selection of the content. (\textit{vi}) Neural networks: More recently, there has been some efforts on utilizing neural networks and sequence-to-sequence models \cite{sutskever2014sequence} for generating summaries of short texts and sentences \cite{rush-chopra-weston:2015:EMNLP,chopra-auli-rush:2016:N16-1}. Most of these works have focused on general domain summarization and news articles. Scientific articles are much different than news articles in elements such as length, language, complexity and structure \cite{Teufel:2002}.

One of the first works in scientific article summarization is done by \cite{Teufel:2002} where the authors trained a supervised Naive Bayes classifier to select informative content for the summary. Later, the impact of citations to generate scientific summaries was realized \cite{elkiss2008blind}. In the work by \cite{Qazvinian2013}, the authors proposed an approach for citation-based summarization based on a clustering approach, while in \cite{abu2011coherent} and \cite{jha2015surveyor}, the focused on producing coherent scientific summaries. We argue that citation texts by themselves are not always accurate and they lack the context of the cited paper. Therefore, if we only use the citation texts for scientific summarization, the resulting summary would potentially suffer from the same problems, and it might not accurately reflect the claims made in the original paper. We address this problem by leveraging the citation contexts from the reference paper. We also utilize the inherent discourse structure of the scientific documents to capture the important content from all sections of the paper.

We present a comprehensive framework for scientific summarization which utilizes and builds upon our earlier efforts \cite{Cohan2015,cohan-goharian:2015:EMNLP,cohan2017contextualizing}. We propose new approaches for citation contextualization. We further extend our experiments on an additional dataset (CL-SciSum 2016) and evaluate our approaches on both TAC and CL-SciSum datasets, providing detailed analysis.


\section{Methodology}
\label{sec:methodology}

Our proposed method is a pipeline for summarizing scientific papers. It consists of the following steps:
\begin{enumerate}
  \item citation contextualization (extracting the relevant context from the reference paper)
  \item identifying the discourse facet of the extracted context
  \item summarization
\end{enumerate}
We first explain our proposed methods for contextualization, we then describe our approach for identifying discourse facets of the citation contexts, and finally we outline our summarization approach.

\subsection{Citation contextualization}
\label{sec:contextualization}

Citation contextualization refers to extracting the relevant context from the reference article for a given citation text. We propose the following three approaches for this problem: (\textit{i}) Query reformulation, (\textit{ii}) Word embeddings and domain knowledge, and (\textit{iii}) Supervised classification.

\subsubsection{Query reformulation (QR)}
\label{subsubsec:qr}

We cast the contextualization problem as an Information Retrieval (IR) task. We first extract textual spans from the reference article and index them using an IR model. The textual spans are of granularity of sentences. In order to capture longer contexts (those consisting of multiple consecutive sentences), we also index sentence n-grams. That is, we index each n consecutive sentences as a separate text span.\footnote{we indexed up to 3 consecutive sentences in our experiments.} After constructing the index, we consider the citation text as the query, and we seek to find the relevant context from the indexed spans. Since the citation texts are often longer than usual queries in standard IR tasks, we apply query reformulation methods on the citation to better retrieve the related context. We utilize both general and domain-specific query reformulations for this purpose. We first remove the citation markers (author names and year, and numbered citations) from the citations, as they do not appear in the reference text and hence are not helpful. We design several regular expressions to capture these names. The proposed query reformulation (QR) methods are described below:

\paragraph{Query reduction}

Since the citation texts are usually more verbose than standard queries, there might be many uninformative terms in them that do not contribute in finding the correct context. Hence, we apply query reduction methods to only retain the important concepts in the citation. After removing the stop words from the citation, we further experiment with the following three query reduction methods:

\begin{enumerate}
  \item Noun phrases (QR-NP). Citation texts are usually linguistically well-formed, as they are extracted from scientific papers. This allows us to apply a variety of linguistic tagging and chunking methods to the query to capture the informative phrases. Previous works have shown that noun phrases are good representation of informative concepts in the query \cite{bendersky2008discovering,huston2010evaluating,hulth2003improved}. We thus extract noun phrases from the citation text and omit all other terms.

  \item Key concepts (QR-KW). Key concepts or keywords are single or multi-word expressions that are informative in finding the relevant context. We use the Inverted Document Frequency (IDF) \cite{sparck1972statistical} measure to find the key concepts. The terms that are prevalent throughout all the text spans do not provide much information in retrieval. IDF values help capturing the terms and concepts that are more specific. For key concept extraction, we limit the IDF values between some threshold that can be tuned according to the dataset.\footnote{We empirically set this threshold to 1.9 and 2.2 for the TAC and CL-SciSum datasets, respectively.} We consider phrases of up to three terms.

  \item Ontology (QR-Domain). Domain-specific ontologies are expert curated lexicons that contain domain-specific concepts. In this reformulation method, we use an ontology to only keep important (domain-specific) concepts in the query. Since the TAC dataset is in the biomedical domain, we use the UMLS \cite{bodenreider2004unified} thesaurus which is a comprehensive ontology of biomedical concepts. We specifically use the SNOMED CT \cite{snomed2011systematized} subset of UMLS.

\end{enumerate}

%

As explained in Section \ref{subsubsec:qr}, the indexing approach also contains consecutive sentences. Therefore, our retrieval approach can find text spans that have overlaps with each other. Furthermore, retrieving multiple spans from around the same location in the text signals the importance of that specific location.
We apply a reranking and merging method to the retrieved spans to remove shared spans and better rank the more relevant context. We merge the two overlapping spans if the retrieval score of the larger span is higher than the smaller span.
We also evaluated other query reformulation methods such as Pseudo Relevance Feedback \cite{cao2008selecting}; however, they performed worse than the baseline and thus we do not discuss them further.

\subsubsection{Contextualization using word embeddings and domain knowledge}

To explicitly account for terminology variations and paraphrasing between the citing and the cited authors, we propose another model for citation contextualization utilizing word embeddings and domain-specific knowledge.

\paragraph{Embeddings.} Word embeddings or distributed representations of words are mapping of words to dense vectors according to a distributional space, with the goal that similar words will be located close to each other \cite{bengio2013representation}. We extend the Language Modeling (LM) for information retrieval model \cite{ponte1998language} by utilizing word embeddings to account for terminology variations. Given a citation text (query) $q$, and a reference span (document) $d$, the LM scores $d$ based on the probability that $d$ has generated $q$ ($p(d|q)$).
Using standard simplifying assumptions of term independence and uniform document prior, we have:

\begin{equation}
  p(d|q) \propto p(q|d) = \prod\limits_{i=1}^{n}p(q_i|d)
\end{equation}

\noindent where $q_i \; (i=1,...,n)$ are the terms in the query. In LM with Dirichlet Smoothing \cite{zhai2004study}, $p(q_i|d)$ is calculated using a smoothed maximum likelihood estimate:

\begin{equation}\label{eq:eq2}
  p(q_i|d) = \frac{f(q_i,d)+\mu\;p(q_i|C)}{\sum_{w\in V} f(w,d) + \mu}
\end{equation}

\noindent where $f$ is the frequency function, $p(q_i|C)$ shows the background probability of term $q_i$ in collection $C$, $V$ is the entire vocabulary, and $\mu$ is the Dirichlet parameter.

Our model extends the above formulation (Eq. \ref{eq:eq2}) by using word embeddings. In particular we estimate the probability $p(q_i|d)$ according to the following equation:


\begin{equation}\label{eq:p1}
p(q_i|d) = \frac{\sum_{dj\in d} s (q_i,d_j)+\mu\;p(q_i|C)}{\sum_{w\in V} \sum_{d_j\in d} s(w,d_j) + \mu}
\end{equation}

\noindent where $d_j$ are terms in the document $d$, and $s$ is a function that captures the similarity between the terms and is defined as:

\begin{equation}\label{eq:simil}
  s(q_i,d_j) =
  \begin{cases}
\mathlarger{\phi}\big(e(q_i),e(d_j)\big), & \text{if } e(q_i).e(d_j) > \tau \\
  0,              & \text{otherwise}
  \end{cases}
\end{equation}

\noindent where $e(q_i)$ shows the unit vector corresponding to the embedding of word $q_i$, $\tau$ is a threshold, and $\phi$ is a transformation function. Below we explain the role of parameter $\tau$ and the transformation function $\phi$.

\begin{table}[t]
\footnotesize
\centering
\setlength{\tabcolsep}{8pt}
\begin{tabular}{@{}llc@{}} \toprule
word 1   & word 2    & Similarity     \\ \midrule
marker   & mint      & 0.11 \\
notebook & sky       & 0.07 \\
capture  & promotion & 0.12 \\ \midrule
blue     & sky       & 0.31 \\
produce  & make      & 0.43 \\ \bottomrule
\end{tabular}
\caption{Example of similarity values between terms according to the dot product of their corresponding embeddings. Using the pre-trained Word2Vec model on Google News corpus. The top part of the table shows pairs of random words, while the bottom part shows similarity values for pairs of related words.}
\label{tab:emb-example}
\end{table}

\begin{figure}
  \centering
  \includegraphics[width=0.7\linewidth]{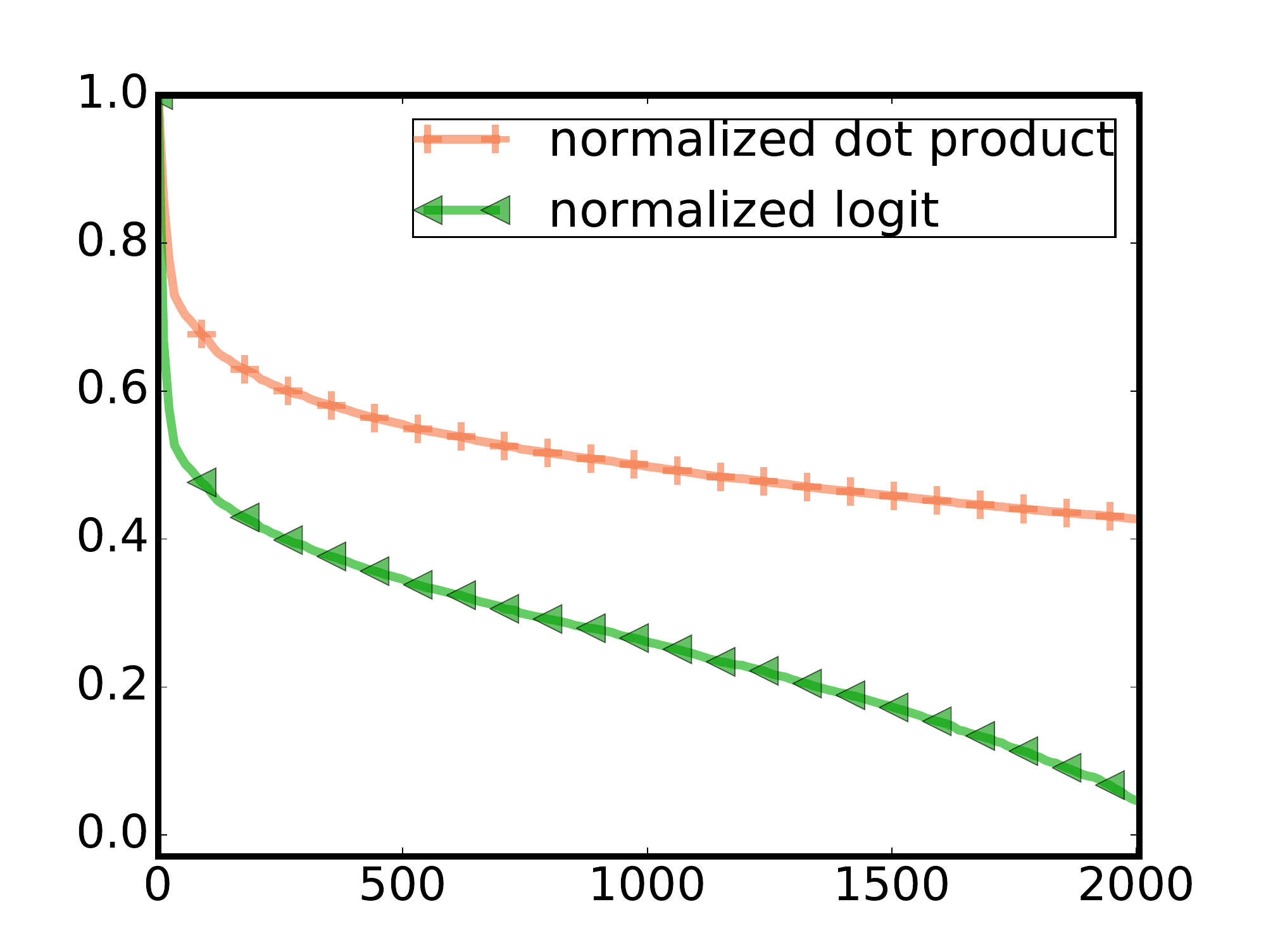}
  \caption{Normalized similarity values between a random word in the embedding model and the top 2000 similar words to it. The x axis is the word indexes and the y axis is the similarity values. The orange line with + markers shows the original similarity values, while the green line with triangle markers shows the transformed values using the logit function. The logit function, dampens the similarity values of less similar words.}
  \label{fig:dampen}
\end{figure}

Word embeddings can capture the similarity values of words according to some distance function. Most embedding methods represent the distance in the distributional semantics space. Therefore, similarities between two words $q_i$ and $d_j$ can be captured using the dot product of their corresponding embeddings (i.e.  $e(q_i).e(d_j)$). While high values of this product suggest syntactic and semantic relatedness between the two terms \cite{mikolov2013distributed,pennington2014glove,hill2015simlex}, many unrelated words have non-zero dot products (an example is shown in Table \ref{tab:emb-example}). Therefore, considering them in the retrieval model introduces noise and hurts the performance. We address this issue by first considering a threshold $\tau$ below which all similarity values are squashed to zero. This ensures that only highly relevant terms contribute to the retrieval model. To identify an appropriate value for $\tau$, we select a random set of words from the embedding model and calculate the average and standard deviation of point-wise absolute values of similarities between the pairs of terms from these samples. We then set $\tau$ to be two standard deviations larger than the average similarities, to only consider very high similarity values. We also observe that for high similarity values between the terms, the values are not discriminative enough between more or less related words. This is illustrated in Figure \ref{fig:dampen} where we can see that the most similar terms to the given term are not too discriminative. In other words, the similarity values decline slowly as moving away from top similar words. We instead want only very top similar words to contribute to the retrieval score. Therefore, we transform the similarity values according to a \textit{logit} function (equation \ref{eq:logit}) to dampen the effect of less similar words (see Figure \ref{fig:dampen}):

\begin{equation}\label{eq:logit}
  \phi(x) = \log(\frac{x}{1-x})
\end{equation}

While any approach for training the word embeddings could be used, we use the Word2Vec \cite{le2014distributed} method, which has proven effective in several word similarity tasks. We train Word2Vec on the recent dump of Wikipedia.\footnote{\url{https://dumps.wikimedia.org/enwiki/}} Since the TAC dataset is in biomedical domain, we also train embeddings on a domain-specific collection; we use the TREC Genomics collections, 2004 and 2006 \cite{hersh2009genomics} which together consist of 1.45 billion tokens.

\paragraph{Incorporating domain knowledge}

Word embedding models learn the relationship between terms by being trained on a large corpus. They are based on the distributional hypothesis \cite{harris1954distributional} which states that similar words appear in similar contexts. While these models have been very successful in capturing semantic relatedness, recent related works have shown that domain ontologies and expert curated lexicons may contain information that are not captured by embeddings \cite{mrkvsic2016counter,hill2015simlex,faruqui2015retrofitting}; hence, we account for the domain knowledge according to the following.

\begin{itemize}
  \item Retrofitting embeddings: In this method, we apply a post-processing step called retrofitting \cite{faruqui2015retrofitting} to the word embeddings used in the model. Retrofitting optimizes an objective function that is based on relationships between words in a lexicon; it intuitively pulls closer the words that are related to each other and pushes farther the words that are not related to each other according to a given ontology. For the ontology, since TAC data is in biomedical domain we use two domain-specific ontologies, \textsc{Mesh}\footnote{MEdical Subject Headings} \cite{lipscomb2000medical} and Protein Ontology (PRO).\footnote{http://pir.georgetown.edu/pro/} For the CL-SciSum data, since it is less domain-specific, we use the WordNet lexicon \cite{miller1995wordnet}.

  \item Interpolating in the LM: In this method, instead of modifying the word vectors, we incorporate the domain knowledge directly in the retrieval model. We do so by interpolation of two following probability estimates:

  \begin{equation}
    p(q_i|d) = \lambda p_1(q_i|d)+(1-\lambda) p_2(q_i|d)
  \end{equation}

  \noindent where $p_1$ is estimated using Eq. \ref{eq:p1} and $p_2$ is a similar model that counts in the \textit{is-synonym} relations ($\issyn$) in calculating similarities. Its formulation is exactly like Eq. \ref{eq:p1} except it replaces the function $s$ with the following function:

  \begin{equation}
    s_2(q_i,d_j){=}
\begin{cases}
1, & \text{if } q_i{=}d_j \\
\gamma, & \text{if } q_i \; \issyn \; d_j \\
0, & \text{o.w.}
\end{cases}
  \end{equation}
\end{itemize}

This function is essentially partially counting the synonyms in calculation of the probability estimate $p(q_i|d)$ by the amount of $\gamma$. We empirically set the value of $\gamma$. Word embedding based methods are shown by WE in short in the results.

\subsubsection{Supervised classification}
\label{subsec:supervised}

The two previous context retrieval models are unsupervised and as such do not take advantage of the already labeled data. CL-SciSum dataset includes separate training and testing sets which allow us to also investigate supervised approaches. We propose a feature-rich classifier to find the correct context for each given citation. Our approach aims to capture the semantic relatedness between a given citation text and a candidate context sentence. We specifically utilize the following features to capture this relatedness:

\begin{itemize}
  \item Word match: counts the number of identical words between the source citation text and the candidate reference context normalized by length.
  \item Fuzzy word match: same as above, with the difference that we use character n-grams to capture partial matches between the words.
  \item Embedding-based alignment: measures the similarity between the source and target sentences using word embedding alignment.
  Specifically for the two sentences $S_1$ and $S_2$, the following function $f$ scores the sentences based on their similarity:
  \begin{equation}
    f(S_1,S_2) = \frac{\sum_{w\in S_1}\max_{v\in S_2} s(w,v)}{|S_1|}
  \end{equation}
  \noindent where $s$ is a similarity function according to the equation \ref{eq:simil}. Intuitively, $f$ captures the similarity between the two sentences without only relying on lexical overlaps; it takes into account the similarity values between the terms.
  \item Distance between average of embeddings: measure the similarity between the two sentences by dot product of the average of their constituent word vectors.
  \item BM25 similarity score \cite{robertson2009probabilistic} between the citation text and the candidate reference.
  \item Tf-idf and count vectorized similarities: dot product between the sparse tf-idf weighted or count weighted vectors associated with the source citation and target reference context.
  \item Character n-gram Tf-idf and count vectorized similarities: same as above, except that we used 3-gram characters to allow partial word matches.
\end{itemize}

We train a standard linear classifier (e.g. Logistic Regression) using these features to identify the correct context for a given citation text.

\subsection{Identifying discourse facets}
\label{sec:discourse}

The organization of scientific papers usually follows a standardized discourse pattern, where the authors first describe the problem or motivation, then they talk about their methods, then the results, and finally discussion and implications. Our goal is to capture the important content from all sections of the paper; therefore, after extracting the citation contexts, we identify the associated discourse facet for each of the citation contexts retrieved from the previous step. Each citation context refers to some specific discourse facets of the reference document. To identify the correct discourse facets, we train a simple supervised model with features listed in Table \ref{tab:facet-feats}. Essentially, we use the citation text and the extracted reference context represented by character n-grams, the verbs in the context sentence, and the relative position of the retrieved context in the paper as features for the classifier. While the textual features (citation and it's context) were the most helpful, we empirically observed slight improvements by incorporating the verb and section position features. We train the model using an SVM classifier \cite{wang2012baselines}. For the textual features, we transform them using character n-grams to allow fuzzy matching between the terms.

\begin{table}[t]
  \footnotesize
  \renewcommand{\arraystretch}{1.2}
  \centering
    \begin{tabular}{@{}l@{}}
    \toprule
     \textbf{Feature Name}                \\ \midrule
    Citation Text               \\
    Extracted Reference Context \\
    Verb Features               \\
    Ralative Section Position   \\ \bottomrule
  \end{tabular}
    \caption{Features for identifying discourse facets.}\label{tab:facet-feats}
\end{table}

\subsection{Generating the summary}

After extracting reference contexts for the citations as described in Section \ref{sec:contextualization}, and identifying their discourse facet (Section \ref{sec:discourse}), we generate a summary of the reference paper.
Our goal is to create a summary that contains information from different discourse facets of the paper. This helps not only in diversifying the content in the summary, but also in creating a more coherent summary. To generate a summary, we first identify the most representative sentences in each group. Intuitively, we only need a few top representative sentences from each discourse facet to include in the summary. In order to find the most representative sentences, we consider sentences in each facet as nodes and their similarities as weighted edges in a graph. We then apply the ``power method'' \cite{erkan2004lexrank} which is an algorithm similar to the PageRank random walk ranking model \cite{page1999pagerank}, that finds the most central nodes in a graph. It works by iteratively updating the score of each sentence according to its centrality (total weight of incoming edges) and the centrality of its neighbors. After ranking the sentences in each group according to their centrality score, we select sentences for the final summary. We use the following methods for creating the final summary:

\begin{itemize}
  \item Iterative. This method simply iterates over the discourse facets and selects the top representative sentence from each group until the summary length threshold is met.
  \item Greedy. The iterative approach could result in similar sentences ending up in the summary; this results in redundant information and potential exclusion of  other important aspects of the paper from the summary. To address this potential problem, we use a heuristic that accounts for both the informativeness of candidate sentence and their novelty with respect to what is already included in the summary. Maximal Marginal Relevance \cite{carbonell1998use} is one such heuristic that has these properties. It is based on the linear interpolation of the informativeness and the novelty of the sentences.
\end{itemize}


\section{Experiments}

\subsection{Data} We conducted our experiments on two scientific summarization datasets. The first dataset is the TAC 2014 scientific summarization dataset.\footnote{http://tac.nist.gov/2014/BiomedSumm/} The TAC benchmark is in biomedical domain and is publicly available upon request from NIST.\footnote{National Institute of Standards and Technology} The second dataset is the 2016 CL-SciSumm dataset \cite{jaidka2016overview} which is available on a public repository\footnote{https://github.com/WING-NUS/scisumm-corpus} and contains scientific articles from the computational linguistics domain. To our knowledge, these two are the only datasets on scientific summarization.

\begin{figure*}[t]
\centering
\hfill
\subfigure[TAC dataset]{\centering \includegraphics[width=0.4\linewidth]{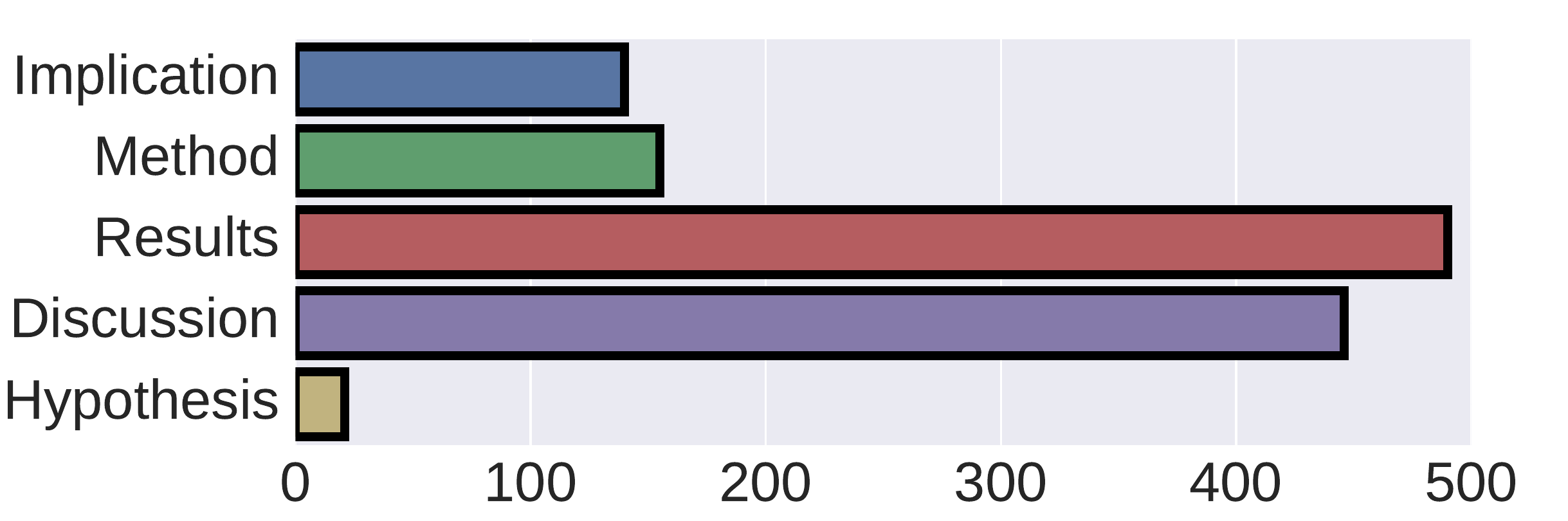}}
\hfill
\subfigure[CL-Scisum dataset]{\centering \includegraphics[width=0.4\linewidth]{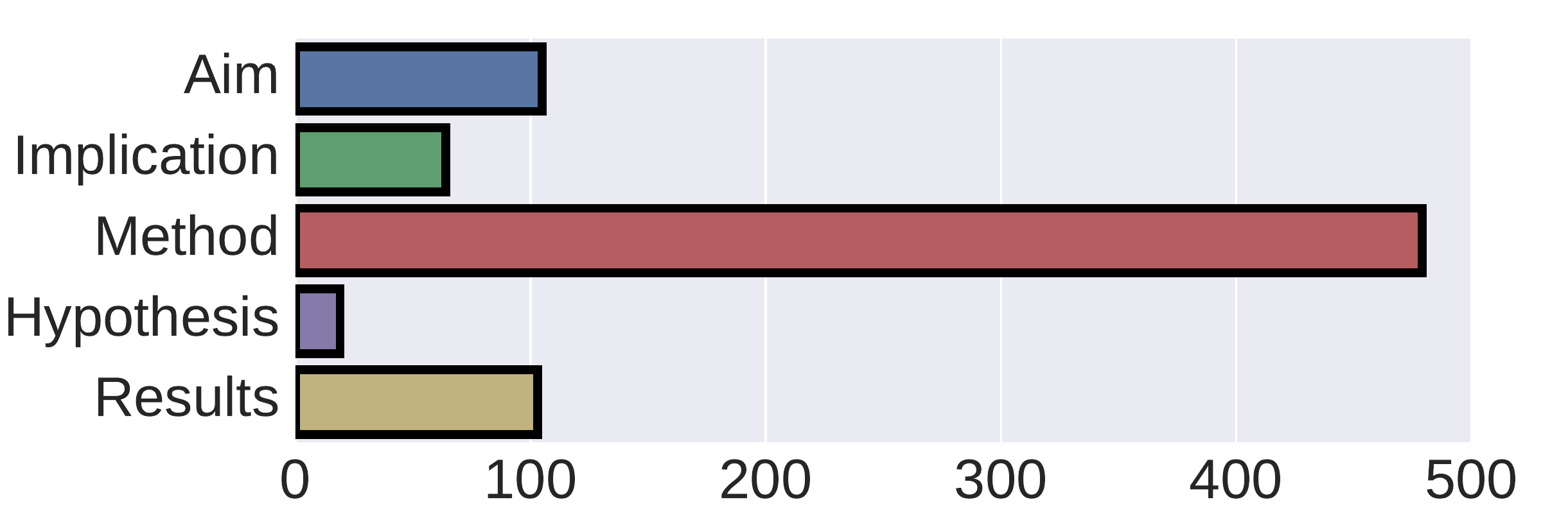}}
\hfill
\caption{Distribution of discourse facets in each dataset.}
\label{fig:data-fact}
\end{figure*}

The TAC dataset only has one training set consisting of 20 topics. There is one reference article in each topic and another set of articles citing the reference. For each topic, 4 annotators have identified the relevant contexts, the correct discourse facet, and they have written a summary. The documents are provided as plain text files and there is no predefined sentence boundaries and sections.
On the other hand, the CL-SciSumm data contain separate train, development, and test sets with 30 topics in total. Similar to TAC, each topic consists of reference and a set of citing articles but in the computational linguistics domain. The articles are in xml format with known sentence boundaries and sections. Another distinction is that topics in the CL-SciSumm data are annotated by one annotator at a time. The full statistics of the datasets is illustrated in Table \ref{tab:data}. The distribution of the discourse facets in the two datasets is also shown in Figure \ref{fig:data-fact}. Since the two datasets are in different domains, the difference between the distribution of the facets is expected.

\begin{table}[t]
  \renewcommand{\arraystretch}{1.2}
  \setlength{\tabcolsep}{3pt}
\centering
\small
\begin{tabular}{@{}lcc@{}}
\toprule
                              Characteristic     & TAC  & CL-SciSum \\ \midrule
\# Documents                        & 220  & 506        \\
\# Reference Documents             & 20   & 30         \\
Avg. \# Citing Docs for each Ref    & 15.5 & 15.9       \\
Total \# Citation Texts               & 313  & 702        \\
Avg. Gold summary length (words)   & 235.6      &   134.2    \\
Stdev. Gold summary length (words) & 31.2         &  27.9     \\
Separate train test sets           & No   & Yes        \\ \bottomrule
\end{tabular}
\caption{Characteristics of the datasets. \#: number of, Avg: average, and Stdev: standard deviation.}
\label{tab:data}
\end{table}

\begin{table*}[t]
\centering
\footnotesize
\renewcommand{\arraystretch}{1.2}
\setlength{\tabcolsep}{5.5pt}
\begin{tabular}{@{}lccccc@{}}
\toprule
 &                             \multicolumn{3}{c}{Character offset overlap}  & \multicolumn{2}{c}{\textsc{Rouge}} \\
 \cmidrule(lr){2-4} \cmidrule(lr){5-6}
  Method                       & $P_{char}$ & $R_{char}$ & $F_{char}$ & \textsc{Rouge}-2 & \textsc{Rouge}-3 \\ \midrule
Baselines \\
                                                                                \rowindent BM25 \cite{robertson2009probabilistic}                         & 19.5       & 18.6       & 17.8       & 23.2             & 16.3             \\
                                                                                \rowindent VSM                          & 20.5       & 24.7       & 21.2       & 26.4             & 20.0             \\
                                                                                \rowindent LMD \cite{zhai2004study}                         & 21.3       & 26.7       & 22.3       & 27.2             & 20.8             \\
                                                                                \rowindent LMD + LDA \cite{jian2016simple}                   & 22.6       & 24.8       & 22.3       & 26.4             & 20.1             \\ \midrule

                                                                                This work \\
                                                                                \rowindent  QR-Domain                    & 24.1       & 23.7       & 21.8       & 25.0             & 20.8             \\
                                                                                \rowindent QR-NP                        & 22.6       & 28.9       & 23.8       & 28.0             & 21.8             \\
                                                                                \rowindent QR-KW                        & 22.6       & 29.4       & 24.1       & 28.2             & 22.2             \\
           \rowindent $\mathrm{WE}_{wiki}$         & 21.8       & 28.5       & 23.2       & 26.9             & 20.9             \\
                                                                                 \rowindent $\mathrm{WE}_{Bio}$          & 23.9       & 31.2       & 25.5      & 29.2              & 23.1            \\
                                                                                \rowindent  $\mathrm{WE}_{Bio+Retrofit}$ & 24.8       & \bf{33.6}  & 26.4      & \bf{30.7}         & 24.0             \\
                                                                                \rowindent $\mathrm{WE}_{Bio}+Domain$  & \bf{25.4}  & 33.0       & \bf{27.0} & 30.6              & \bf{24.4}            \\ \bottomrule
\end{tabular}
\caption{Results of citation contextualization on TAC 2014 dataset. The reported results are based on top 10 retrieved contexts. The top part shows the baselines and the bottom part shows our proposed model. Values are percentages. QR-Domain: Query Reformulation by Domain Ontology (UMLS), QR-NP: Query Reformulation by Noun Phrases, QR-KW: Query Reformulation by Key Words,  $\mathrm{WE}_{wiki}$: Word Embedding model with Wikipedia embeddings, $\mathrm{WE}_{Bio}$: Word Embedding model with biomedical embeddings, $\mathrm{WE}_{Bio+Retrofit}$: Incorporating domain knowledge in biomedical embeddings by retrofitting, $\mathrm{WE}_{Bio}+Domain$: Interpolated language model. }
\label{tab:task1a-tac}
\vspace{-12pt}
\end{table*}

\subsection{Citation contextualization}
\label{subsc{res:context}}

\paragraph{Evaluation} Evaluation of the retrieved contexts is based on the overlap of the position of the retrieved contexts and the gold standard contexts. Per TAC guidelines\footnote{http://tac.nist.gov/2014/BiomedSumm/guidelines.html}, evaluation of the TAC benchmark was performed using character offset overlaps weighted by human annotators. More formally, for a set of system retrieved contexts $S$, and gold standard context $R=\{R_1\cup R_2\cup....\cup R_m\}$ by $m$ annotators, the weighted character based precision ($P_{char}$) and recall ($R_{char}$) are defined as follows:

\noindent\begin{minipage}{.47\linewidth}
\small
\begin{equation}
  P_{char} = \frac{\sum_{i}^m|S\cap R_i|}{m \times |S|}
\end{equation}
\end{minipage}%
~~~~~
\begin{minipage}{.47\linewidth}
  \small
  \begin{equation}
    R_{char} = \frac{\sum_{i}^m|S\cap R_i|}{\sum_i^m|R_i|}
  \end{equation}
\end{minipage}

The official metric for the CL-SciSum challenge was sentence level overlaps of the retrieved contexts with the gold standard. This was possible because unlike the articles in TAC which were in plaintext format, the sentence boundaries in CL-SciSum were pre-specified. We also report character level metrics for the CL-SciSum corpus; as we will see, the character level and sentence level metrics are more or less comparable.

One problem with position based evaluation metrics (character, or sentence) is that a system might retrieve a context that is in a different position than gold standard, but similar to the content of the gold standard. In such cases, the system is not rewarded at all. This is possible because authors might talk about a similar concept in different sections of the paper. To consider textual similarities of the retrieved context with the gold standard, we also compute \textsc{Rouge}-N scores \cite{lin2004rouge}.

\paragraph{Comparison}

To our knowledge, no review paper about the TAC challenge was released. Hence, for the TAC dataset, we compare our method against the following baselines:

\begin{itemize}
  \item VSM. Ranking by Vector Space Model (VSM) with tf-idf weighting of the citations and the target reference contexts.
  \item BM25. BM25 scoring model \cite{jones2000probabilistic} which is a probabilistic framework for ranking the relevant documents based on the query terms appearing in each document, regardless of their relative proximity.
  \item LMD. Language modeling with Dirichlet smoothing (LMD) \cite{zhai2004study} is a probabilistic framework that models the probability of documents generating the given query.
  \item LMD-LDA. An extension of the LMD retrieval model using Latent Dirichlet Allocation (LDA) which is recently proposed \cite{jian2016simple}. This model considers latent topics in ranking the relevant documents
\end{itemize}

For the CL-SciSum data, we also compare against the top 5 best performing system. For brief description about these approaches refer to section \ref{sec:related}.

\begin{table*}[t]
\centering
\small
\renewcommand{\arraystretch}{1.2}
\setlength{\tabcolsep}{4pt}
\begin{tabular}{@{}lcccccccc@{}}
\toprule
                                       & \multicolumn{3}{c}{Sentence overlap} & \multicolumn{2}{c}{\textsc{Rouge}}  & \multicolumn{3}{c}{Character offset overlap}  \\
 \cmidrule(lr){2-4} \cmidrule(lr){5-6} \cmidrule(lr){7-9}
  Method                                      & $P_{sent}$ & $R_{sent}$ & $F_{sent}$ & \textsc{Rouge}-2 & \textsc{Rouge}-3 & $P_{char}$ & $R_{char}$ & $F_{char}$ \\ \midrule
                                                                      Other methods\\
                                                                      \rowindent BM25 \cite{robertson2009probabilistic}      & 8.2        & 18.0       & 10.5       & 15.2             & 13.0             & 9.0        & 19.9       & 11.8       \\
                                                                      \rowindent VSM                                         & 8.3        & 22.3       & 11.6       & 14.8             & 12.7             & 8.5        & 25.7       & 12.1       \\
                                                                      \rowindent LM  \cite{zhai2004study}                    & 7.9        & 24.8       & 11.6       & 14.3             & 12.6             & 8.4        & \bf{26.1}       & 12.2       \\
                                                                      \rowindent TSR \cite{klampfl2016identifying}           & 5.3        & 4.7        & 5.0        & -                & -                & -          & -          & -          \\
                                                                      \rowindent Tf-idf + Neural Net \cite{Nomoto2016NEALAN} & 9.2        & 11.1       & 10.0       & -                & -                & -          & -          & -          \\
                                                                      \rowindent SVM Rank \cite{Cao2016PolyUAC}              & 8.8        & 13.1       & 10.3       & -                & -                & -          & -          & -          \\
                                                                      \rowindent Jaccard Fusion \cite{Li2016CISTSF}          & 8.3        & \bf{26.1}  & 12.5       & -                & -                & -          & -          & -          \\
                                                                      \rowindent Tf-idf+stem \cite{moraes2016university}     & 9.6        & 22.4       & 13.4       & -                & -                & -          & -          & -          \\       \midrule
This work\\
                                                                      \rowindent QR-NP                                       & 8.8        & 20.4       & 12.2       & 15.8             & 13.6             & 9.7        & 23.8       & 13.2       \\
                                                                      \rowindent QR-KW                                       & 9.0        & 21.3       & 12.6       & 16.0        & 13.8             & 9.6        & 23.3  & 13.0       \\
                                                                      \rowindent WE$_{wiki}$                                 & 9.8        & 24.1       & \bf{13.9}  & 14.5             & 12.5             & 9.4        & 22.1       & 12.5       \\
                                                                      \rowindent WE$_{wiki+Retrofit}$                           & 9.8        & 23.8       & 13.8       & 14.7             & 13.6             & 8.2        & 22.3       & 12.0       \\
                                                                      \rowindent Supervised                                  & \bf{11.3}  & 17.8       & 13.7       & \bf{17.5}        & \bf{15.0}        & \bf{12.0}  & 17.8       & \bf{13.7}  \\ \bottomrule
\end{tabular}
\caption{Results of citation contextualization on CL-SciSum 2016 dataset. The reported values are percentages. The top part shows the baselines and state of the art models, while the bottom part shows our methods. P: Precision, R: Recall, F:F1-score. ``sent'' subscript shows overlap by sentences and ``char'' subscript shows character offset overlaps. QR-NP: Query Reformulation by Noun Phrases, QR-KW: Query Reformulation by Key Words, $\mathrm{WE}_{wiki}$: Word Embedding model with Wikipedia embeddings, $\mathrm{WE}_{wiki+Retrofit}$: Incorporating domain knowledge in embeddings by retrofitting}
\label{tab:task1a-cl}
\end{table*}

\paragraph{Results.}
The results on the TAC dataset are presented in Table \ref{tab:task1a-tac}. We observe that our proposed methods improve over all the baselines. Query Reformulation methods (NP and KW, respectively,) obtain character offset F1-scores of 23.8 and 24.1, which improve the best baseline by 7\% and 8\%. They also obtain higher \textsc{Rouge} scores. This shows that noun phrases and key words can capture informative concepts in the citation that help better retrieving the related reference context. Our models based on word embeddings are also outperforming the baselines in virtually all metrics. General domain embeddings trained on Wikipedia (WE$_{wiki}$) and domain-specific embeddings trained on Genomics data (WE$_{Bio}$), achieve F1-scores of 23.2 and 25.5 with 4\% and 14\% improvement over the best baseline, respectively. Higher performance of the biomedical embeddings in comparison with general embeddings is expected because the words are captured in their correct context. An example is shown in Table \ref{tab:example:words}, where the top similar words to the word ``expression'' are shown. The word ``expression'' in the biomedical context is defined as ``the process by which genetic instructions are used to synthesize gene products''. As we can see, using general domain embeddings, we might fail to capture this notion. Incorporating domain knowledge in the model results in further improvement as shown in last two rows of Table \ref{tab:task1a-tac}. The model using retrofitting WE$_{Bio+Retrofit}$ improves the best baseline by 18\% while the interpolated model (WE$_{Bio}$+Domain) achieves the highest improvement by 21\%. These results show the effectiveness of domain knowledge in the model.

\begin{table}[t]
  \small
\centering
\renewcommand{\arraystretch}{1.2}
\setlength{\tabcolsep}{12pt}
\begin{tabular}{@{}ll@{}}
\toprule
\begin{tabular}[l]{@{}l@{}}General\\  (Wiki)\end{tabular} & \begin{tabular}[l]{@{}l@{}}Domain-specific\\ (Bio)\end{tabular} \\ \midrule
interpretation                                            & upregulation                                                    \\
sense                                                     & mrna                                                            \\
emotion                                                   & protein                                                         \\
function                                                  & induction                                                       \\
show                                                      & cell                                                            \\ \bottomrule
\end{tabular}
\caption{The words with highest similarity values to ``expression'' according to Word2Vec trained on Wikipedia (general domain) and Genomics collections (biomedical domain).}
\label{tab:example:words}
\end{table}

Table \ref{tab:task1a-cl} shows the results for the CL-SciSum dataset. The first 3 rows are baselines that also are reported in TAC evaluation; in addition to those baselines, we also consider top performing state-of-the-art systems of 2016 CL-SciSum (lines 4-8) as additional baselines to compare with. For the CL-SciSum participating systems, we report the official sentence based evaluation metrics; the \textsc{Rouge} scores and character based metrics were not reported in the official evaluation of the task. Some of our methods are specific to the biomedical domain such as WE$_{Bio}$; therefore, we do not evaluate those on the CL-SciSum dataset which is in a completely different domain.

\begin{table}[]
\centering
\footnotesize
\renewcommand{\arraystretch}{1.2}
\begin{tabular}{@{}ll@{}}
\toprule
Feature                            & weight \\ \midrule
character n-gram tf-idf similarity & 0.271  \\
tf-idf similarity                  & 0.201  \\
embedding based alignment          & 0.189  \\
distance average embeddings        & 0.106  \\
bm25 similarity score              & 0.066 \\
character n-gram count similarity  & 0.035 \\
fuzzy word match                   & 0.024  \\
count based similarity             & 0.015 \\
word match                         & 0.013  \\
\bottomrule
\end{tabular}
\caption{The weights (normalized) corresponding to the top features in the supervised method for citation contextualization (CL-SciSum dataset). Tf-idf similarity based features and embedding based features are the most helpful while the count based similarity and word matching features are among the least helpful features.}
\label{tab:feat-weight}
\end{table}

\begin{figure*}[tb]
\centering
\hfill
\subfigure[Effect of the parameter $\gamma$ on the interpolated contextualization model (tuned on the TAC dataset).]{\includegraphics[width=4cm]{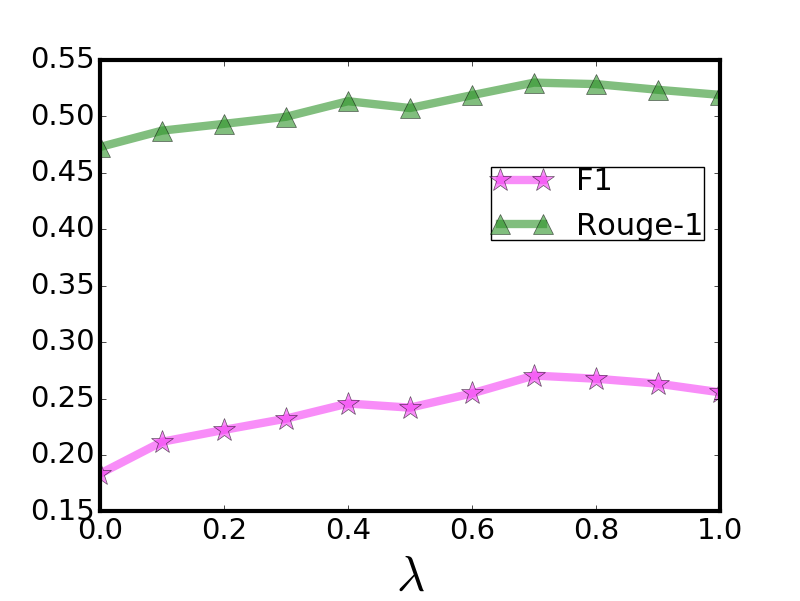}}
\hfill
\subfigure[Effect of the parameter $\lambda$ on the interpolated contextualization model (tuned on the TAC dataset).]{\includegraphics[width=4cm]{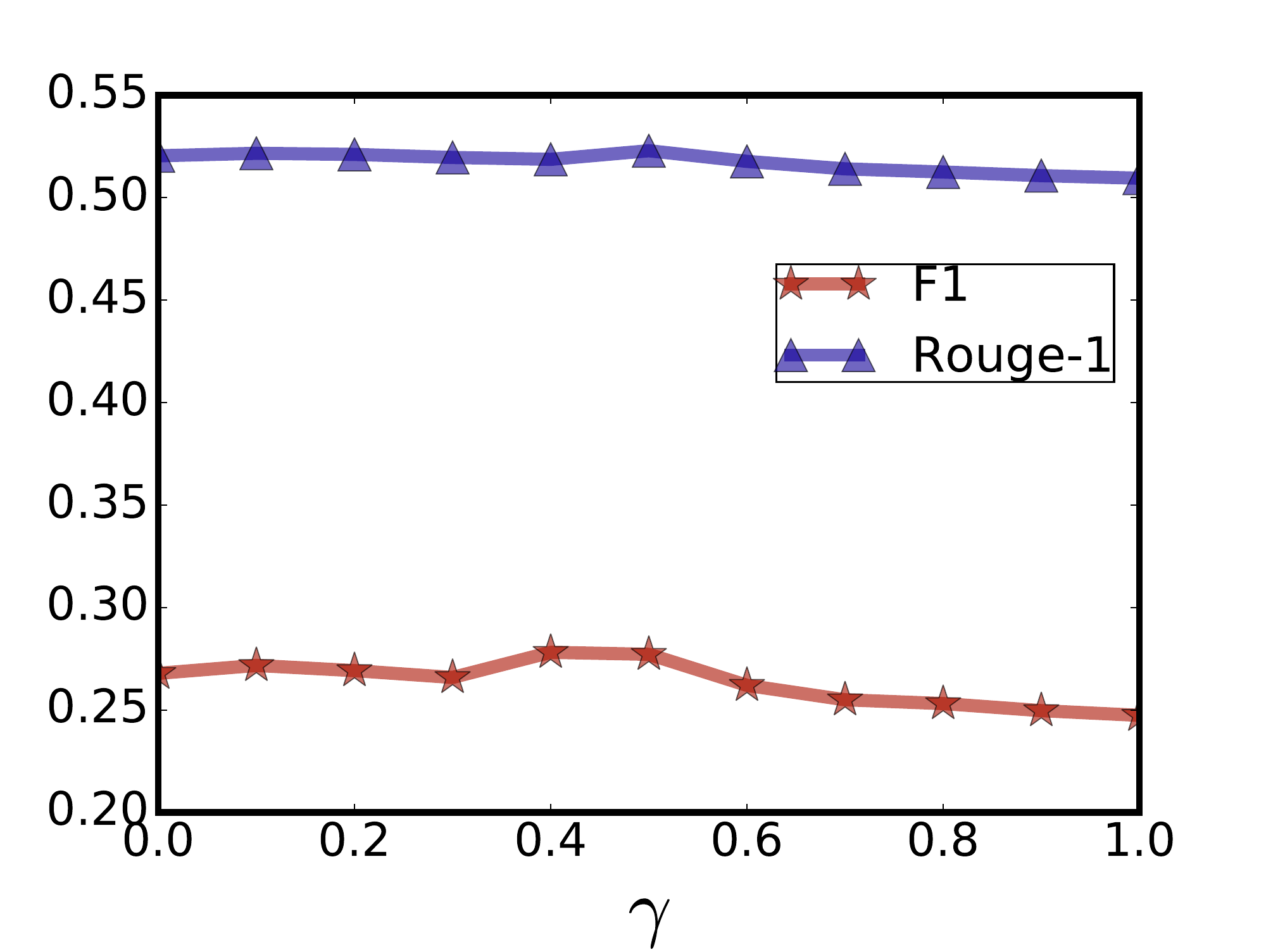}}
\hfill
\subfigure[Effect of the cut-off point in returning the top results for the WE$_{wiki}$ model (tuned on CL-SciSum dataset).\label{subfig1}]{\includegraphics[width=4cm]{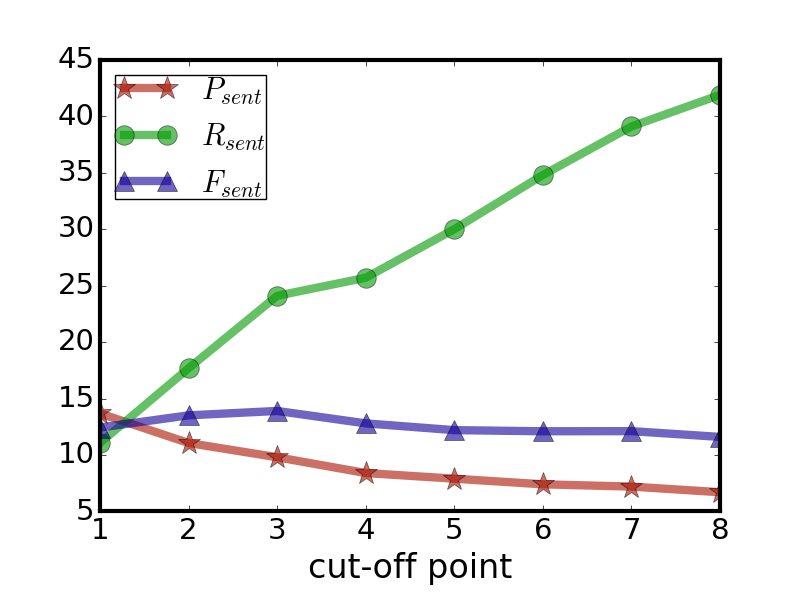}}
\hfill
\caption{Parameters of the model for contextualization.}
\label{fig:params}
\end{figure*}

 As shown in Table \ref{tab:task1a-cl}, our methods outperform the state-of-the-art on this dataset as well. The embedding-based model with Wikipedia trained embeddings (WE$_{wiki}$) achieves the best results with 13.9\% F-1 score of sentence overlaps which is slightly higher than the F-1 score of 13.4 achieved by the best previous work (Tf-idf+stem in the Table) \cite{moraes2016university}. Interestingly, we  observe that retrofitting (WE$_{wiki+Retrofit}$) does not improve over the standard embedding-based approach. This is likely due to the choice of the Wordnet lexicon for retrofitting. While Wordnet contains general domain terms, it does not necessarily capture relationships of words in the context of computational linguistics. In contrast to TAC where we had a domain specific lexicon suitable for the dataset, for the CL-SciSum data we did not find any lexicon capturing the term relationships in the computational linguistics domain. We believe that retrofitting with such lexicon, could result in further improvements. While query reformulation-based approaches improve over most of the baselines, their performance fall below the best baseline system. On the other hand, our supervised method also improves the best baseline, achieving the highest overall prevision (11.3\%) and \textsc{Rouge}-2 (17.5\%) and \textsc{Rouge}-3 scores (15.0\%).\footnote{We do not report results of supervised model on TAC dataset because the TAC data do not have separate train and test sets.} It is encouraging that our embedding-based models (method names starting with ``WE'' in the Table \ref{tab:task1a-cl}), which are unsupervised models achieve the best results on this task and surpass the performance of the feature-rich supervised models. Table \ref{tab:feat-weight} shows the importance of each feature for our supervised method (explained in \S ~\ref{subsec:supervised}). While the most important features are n-gram and character n-gram based tf-idf similarity, embedding based alignment and distance of average embeddings are also important in finding the correct context.

 \begin{table}[t]
   \small
 \centering
 \renewcommand{\arraystretch}{1.2}
 \begin{tabular}{@{}cc@{}}
 \toprule
 Number of Citations &  \begin{tabular}[c]{@{}c@{}} Number of Annotators with at\\  least partial agreement\end{tabular}\\ \midrule
 68          & 4                                            \\
 66          & 3                                            \\
 121         & 2                                            \\
 11          & No agreement                                 \\ \bottomrule
 \end{tabular}
 \caption{The table shows the number of citations grouped by the number of annotators that agree at least partially on the context.}
 \label{tab:agree}
 \end{table}

   \begin{table*}[]
  \small
  \renewcommand{\arraystretch}{1.2}
   \centering
   \begin{tabular}{@{}lccc@{}} \toprule
     Method                                     & P         & R         & F         \\ \midrule
                                                                       Other methods\\
                                                                         \rowindent SMO \cite{saggion2016trainable}            & 35.6      & 3.6       & 6.5       \\
                                                                         \rowindent Decision tree \cite{Cao2016PolyUAC}        & 59.7      & 9.0       & 15.3     \\
                                                                         \rowindent Fusion method \cite{Li2016CISTSF}          & 52.8      & 22.4      & 29.6      \\
                                                                         \rowindent Jaccard cascade \cite{Li2016CISTSF}        & 58.2      & 17.1     & 25.5      \\
                                                                         \rowindent Jaccard Focused Method \cite{Li2016CISTSF} & 57.8      & 22.8      & 31.1     \\ \midrule
   This work \\
                                                                         \rowindent QR-NP                                      & 76.3      & 19.1      & 29.7      \\
                                                                         \rowindent QR-KW                                      & 78.7      & 21.9      & 33.3      \\
                                                                         \rowindent WE$_{wiki}$                                & 82.7     & 22.4      & 33.1      \\
                                                                         \rowindent WE$_{wiki+retro}$                          & 81.7     & 23.4      & 34.8      \\
                                                                         \rowindent Supervised                                 & \bf{83.1} & \bf{23.7} & \bf{36.1} \\ \bottomrule
   \end{tabular}
   \caption{Results for identifying the discourse facets for the retrieved contexts. The metrics are Precision (P), Recall (R), and F1-score (F) of the identified discourse facets contingent on the correct retrieved span.}
   \label{tab:task1b}
 \end{table*}

 As evident from tables \ref{tab:task1a-tac} and \ref{tab:task1a-cl}, the absolute system performances are not high, which further shows that this task is challenging. Since the TAC data are annotated by 4 people, we investigate the difficulty of this task for the human annotators. To do so, we calculate the agreement of the annotators with respect to the relevant context for the citations. Table \ref{tab:agree} shows the number of citations grouped by the number of annotators that agree at least partially on the correct context. As illustrated, there are 68 citations out of 313 that all 4 annotators have partial agreement on the context span. This shows that the contextualization task is not trivial even for the human expert annotators.

 \paragraph{Parameters}
 Our interpolated model of embeddings and domain knowledge (WE$_{Bio}$+Domain) has two main parameters $\gamma$ and $\lambda$. Figure \ref{fig:params} shows the sensitivity of our model to different parameters. We observe that the best performance is achieved when $\gamma=0.8$ and $\lambda=0.5$. Our models retrieve a ranked list of contexts for the citations; we choose a cut-off point for returning the final results. Figure \ref{subfig1} shows the effect of the cut-off point on one of our models.\footnote{The cut-off point has similar effect on all the models.} We observe that the optimal cut-off point for best sentence F1-score is 3.

 \subsection{Identifying discourse facets}

 \paragraph{Evaluation}
 The official metric for evaluation of discourse facet identification is the Precision, Recall and F1-scores of the discourse facets, conditioned on the correctness of the retrieved reference context \cite{jaidka2016overview}. Therefore, we report the results for the CL-SciSum data based on this metric.
 For the TAC dataset, the official metric is the classification accuracy weighted by the annotator agreements.\footnote{http://tac.nist.gov/2014/BiomedSumm/guidelines/} The accuracy for a system returned discourse facet is the number of annotators agreeing with that discourse facet divided by total number of annotators.

 \paragraph{Results}
 Table \ref{tab:task1b} shows the results of our methods compared with the top performing official submitted runs to the CL-SciSum 2016. We do not report the results of low performing systems. The classification algorithm for identifying the discourse facets is the method described in Section \ref{sec:discourse} across all our methods. However, since only the correct retrieved contexts are rewarded, the performance of each model differs based on the accuracy of retrieving the correct contexts. We observe that most of our methods (except for the QR-NP) improve over all the baselines in terms of all metrics. We obtain substantial improvements especially in terms of precision. The best method for identifying the discourse facets is the supervised method (indicated with ``supervised'' in the Table) which obtains 36.1\% F-1 score, improving the best baseline (``Jaccard Focused Method'') by 16\%. Embedding methods also perform well by obtaining F-1 scores of 33.1\% for the Wikipedia embeddings, and 34.8\% for the retrofitted embeddings. These results further show the effectiveness of our contextualization methods along with the proposed classifier for identifying the facets.

 \begin{table}[t]
 \centering
 \renewcommand{\arraystretch}{1.2}
 \footnotesize
 \begin{tabular}{@{}lcccc@{}} \toprule
 Discourse  Facet               & P    & R    & F    & \#   \\ \midrule
 Aim             & 0.93 & 0.36 & 0.52 & 36  \\
 Hypothesis      & 1.00 & 0.20 & 0.33 & 10  \\
 Implication     & 0.85 & 0.26 & 0.39 & 43  \\
 Method          & 0.79 & 0.98 & 0.87 & 250 \\
 Results         & 0.85 & 0.38 & 0.52 & 45  \\ \midrule
 Average/Total & 0.82 & 0.75 & 0.73 & 384 \\ \bottomrule
 \end{tabular}
 \caption{The classifier's intrinsic performance for identifying the discourse facets on the CL-SciSum dataset.}
 \label{tab:facet-cl}
 \end{table}

We also demonstrate the intrinsic performance of our classifier for identifying the discourse facets in Table \ref{tab:facet-cl}. As illustrated, the weighed average F1 performance over all discourse facets is 0.73. One challenge in identifying the discourse facets is the unbalanced dataset and the limited number of training examples for some specific facets. As also reflected in the table, we observe that for categories with smaller number of instances, the performance is generally lower. We therefore believe that having more training samples in the rare categories could further increase the performance.

Table \ref{tab:facet-tac} shows the results of facet identification in the TAC dataset as well as the effect of learning algorithms. Since for the TAC dataset there are 4 annotators, and the official metric is weighted accuracy scores, we also calculate the oracle score by always predicting what the majority of the annotators agree on. The oracle achieves 0.67 percent, suggesting that identifying discourse facets is not trivial for humans. We can see that the SVM classifier achieves the highest results with 81\% relative accuracy to the oracle. For the CL-SciSum dataset, there is only one annotator per discourse facet and therefore, the weighted accuracy metrics translates to simple accuracy scores.

\begin{table}[t]
  \small
\centering
\renewcommand{\arraystretch}{1.2}
\begin{tabular}{@{}lcccc@{}}
\toprule
          & SVM  & RF   & LR   & Oracle \\ \midrule
TAC       & 0.53 & 0.49 & 0.51 & 0.67   \\
CL-SciSum & 0.67 & 0.64 & 0.66 & -      \\ \bottomrule
\end{tabular}
\caption{Effect of learning algorithms in identifying the discourse facets. SVM: Support Vector Machine with Linear Kernel, RF: Random Forest, LR: Logistic Regression, Oracle: Highest achievable score. Numbers are weighted accuracy scores by annotators.}
\label{tab:facet-tac}
\end{table}

\begin{table*}[t]
  \small
\centering
\renewcommand{\arraystretch}{1.2}
\begin{tabular}{@{}lccc@{}} \toprule
                                  & \textsc{Rouge}-2 & \textsc{Rouge}-3 & \textsc{Rouge}-SU4 \\ \midrule
LexRank \cite{erkan2004}                 & 11.8     & 8.1     & 11.4    \\
CLexRank \cite{qazvinian2008scientific}                         & 5.7     & 3.3     & 8.9    \\
SumBasic \cite{vanderwende2007beyond}                         & 8.5     & 3.8     & 11.5    \\ \midrule
SUMMA \cite{saggion2016trainable} & 13.4    &  -       & 9.2     \\
LMKL \cite{conroy2015vector}      & 19.0    &  -       & 11.1    \\
LMeq \cite{conroy2015vector}      & 18.9    &  -       & 12.4    \\
CIST \cite{Li2016CISTSF}          & 21.9    &  -       & 13.6    \\ \midrule
QR-KW-iter                        & 27.6    & 21.4    & 23.4    \\
QR-KW-greedy                      & 28.9    & 22.5    & 24.9    \\
QR-NP-iter                        & 23.0    & 20.9    & 22.6    \\
QR-NP-greedy                      & \bf{30.2}    & \bf{23.9}    & \bf{25.7} \\
WE$_{wiki}$-iter                    & 22.4    & 15.9    & 21.7    \\
WE$_{wiki}$-greedy                  & 23.6    & 18.0    & 20.1    \\
supervised-iter                   & 24.1    & 18.5    & 20.8    \\
supervised-greedy                 & 23.6    & 18.3    & 19.6    \\ \bottomrule

\end{tabular}
\caption{Summarization results on the CL-SciSum dataset. Metrics are \textsc{Rouge} F-scores. The top part shows the baselines and the state-of-the-art systems.
Bottom systems show our method variants based on different contextualization approaches and sentence selection strategy from the discourse facets. \textit{iter} (iterative) and \textit{greedy} refer to the sentence selection approach for the final summary.}
\label{tab:summ-CL}
\end{table*}

\begin{table*}[t]
  \small
\centering
\renewcommand{\arraystretch}{1.2}
\begin{tabular}{@{}lccc@{}} \toprule
                           & \textsc{Rouge}-2 & \textsc{Rouge}-3 & \textsc{Rouge}-SU4 \\ \midrule
LexRank \cite{erkan2004}                 & 12.8     & 5.0     & 17.5    \\
CLexRank \cite{qazvinian2008scientific}                 & 8.9     & 3.9     & 8.3    \\
SumBasic \cite{vanderwende2007beyond}                 & 8.3     & 4.2     & 12.5    \\ \midrule
QR-NP                      & \bf{15.8}    & \bf{6.9}     & 20.4    \\
QR-Domain                  & 13.2    & 5.2     & 18.1    \\
QR-KW                      & 15.0    & 6.6     & 19.8 \\
WE$_{wiki}$                & 13.3    & 5.5     & 17.8    \\
WE$_{Bio}$                 & 13.1    & 4.9     & 18.0    \\
WE$_{Bio+Retrofit}$ & 14.4    & 5.7     & 19.5    \\
WE$_{Bio}$+Domain           & 13.4    & 5.9     & \bf{20.7}    \\ \bottomrule

\end{tabular}
\caption{Summarization results on the TAC dataset. Metrics are \textsc{Rouge} F-scores. The top part shows the baselines and the state-of-the-art systems.
Bottom systems show our method variants based on different contextualization approaches and the greedy sentence selection strategy.}
\label{tab:summ-TAC}
\end{table*}

\begin{figure*}[]
\centering
\scriptsize
\begin{tabular}{p{0.08\textwidth}p{0.85\textwidth}}
\toprule
                                                         & Example summary                                                                                                                                                                                                                                                                                                                                                                                                                                                                                                                                                                                                                                                                                                                                                                                                                                                                                                                                                                                                                                                                                                                                                                                                                                                                                                                                                    \\ \hline
\begin{tabular}[c]{@{}l@{}}Human \\ Summary\end{tabular} & The limited coverage of lexical-semantic resources is a significant problem for NLP systems which can be alleviated by automatically classifying the unknown words. Supersense tagging assigns unknown nouns one of 26 broad semantic categories used by lexicographers to organise their manual insertion into WORDNET. Lexical-semantic resources have been applied successful to a wide range of Natural Language Processing (NLP) problems ranging from collocation extraction and class-based smoothing, to text classification and question answering. Some specialist topics are better covered in WORDNET than others. A considerable amount of research addresses structurally and statistically manipulating the hierarchy of WORDNET and the construction of new wordnet using the concept structure from English. Ciaramita and Johnson,implement a supersense tagger based on the multi-class preceptor classifier, which uses the standard collocation, spelling and syntactic features common in WSD and named entity recognition systems. The authors demonstrate the use of a very efficient shallow NLP pipeline to process a massive corpus. Such a corpus is needed to acquire reliable contextual information for the often very rare nouns they are attempting to supersense tag.                                                            \\ \midrule
\begin{tabular}[c]{@{}l@{}}Our\\ System\end{tabular}     & The limited coverage of lexical-semantic resources is a significant problem for NLP systems which can be alleviated by automatically classifying the unknown words. Ciaramita and Johnson (2003) present a tagger which uses synonym set glosses as annotated training examples. Our approach uses voting across the known supersenses of automatically extracted synonyms, to select a super- sense for the unknown nouns. The extracted synonyms are filtered before contributing to the vote with their supersense(s). Our development experiments are performed on the WORDNET 1.6 test set with one final run on the WORD- NET 1.7.1 test set. In particular, WORDNET (Fellbaum, 1998) has significantly influenced research in NLP. These results also support Ciaramita and Johnson’s view that abstract concepts like communication, cognition and state are much harder. Lexicographers cannot possibly keep pace with language evolution: sense distinctions are continually made and merged, words are coined or become obsolete, and technical terms migrate into the vernacular.                                                                                                                                                                                                                                                                      \\ \midrule
CLexRank                                                 & Another related task is supersense tagging (Ciaramita and Johnson, 2003; Curran, 2005; Ciaramita and Altun, 2006). Supersense tagging (Ciaramita and Johnson, 2003; Curran, 2005) evaluates a model's ability to cluster words by their semantics. In contrast, some research have been focused on using predefined sets of sense-groupings for learning class-based classifiers for WSD. Although we could adapt our method for use with an automatically induced inventory, our method which uses WordNet might also be combined with one that can automatically find new senses from text and then relate these to WordNet synsets, as Ciaramita and Johnson (2003) and Curran (2005) do with unknown nouns.. An additional potential is to integrate automatically acquired relationships with the information found in WordNet, which seems to suffer from several serious limitations (Curran 2005), and typically overlaps to a rather limited extent with the output of automatic acquisition methods. Previous work on prediction at the supersense level (Ciaramita and Johnson, 2003; Curran, 2005) has focused on lexical acquisition (nouns exclusively), thus aiming at word type classification rather than tagging. \\ \bottomrule
\end{tabular}
\caption{Example summary generated by our system (QR-NP-Greedy) on one of the papers from the CL-SciSum dataset, compared with a human written summary and the output generated by CLexRank.}
\label{fig:example-summaries}
\end{figure*}

\subsection{Summarization}

We evaluate our summarization approach against the gold standard summaries written by human annotators. We set the summary length threshold to the average length of summary by words in each dataset (see Table \ref{tab:data}). Table \ref{tab:summ-CL} shows the results for the summarization task. The first lines show the baselines which are existing summarization approaches including the SumBasic \cite{vanderwende2007beyond} algorithm and the original citation-based summarization approach \cite{qazvinian2008scientific}. The next four lines are the top state-of-the-art systems on the CL-SciSum dataset. For the CL-SciSum systems, the official reported results only included \textsc{Rouge}-2 and \textsc{Rouge}-SU4 scores. As illustrated in the table, virtually all our methods improve over the state-of-the-art, showing the effectiveness of our proposed summarization approach. Our best method (QR-NP-greedy) is based on the noun phrases query reformulation using the greedy strategy of sentence selection . It achieves \textsc{Rouge}-2 score of 30.2, which improves over the best baseline by 37.4\%. In general, we can see that the greedy sentence selection strategy works better than the iterative approach. This is because the greedy strategy takes into account both the informativeness and the redundancy of the selected sentences.

Table \ref{tab:summ-TAC} shows the results of summarization using on the TAC dataset. The reported approaches all use the greedy sentence selection strategy as it consistently outperforms the iterative approach. In general, while all our approaches outperform the baseline, query reformulation based approaches achieve the highest \textsc{Rouge} scores; query reformulation method using noun phrases (QR-NP) achieves 15.8 and 6.9 \textsc{Rouge}-2 and \textsc{Rouge}-3 scores, respectively which is the highest scores. The interpolated word embedding based model (WE$_{Bio}$+Domain) achieves the highest \textsc{Rouge-su}4 score (20.7).
Comparing Tables \ref{tab:summ-CL} and \ref{tab:summ-TAC} we notice that the scores for the TAC dataset are lower than that of CL-SciSum. This is due to the length of the generated summaries. As shown in Table \ref{tab:data}, the average human summary length in the TAC data is almost 100 words more than the CL-SciSum summaries. An interesting observation in these two tables is regarding the relative poor performance of the citation-based summarization baseline (CLexRank) that only uses citation texts in comparison with our methods that also take advantage of the citation context and the discourse structure of the articles. This observation further confirms our initial hypothesis that relying only on the citation texts could result in summaries that do not accurately reflect the content of the original paper, and that adding citation contexts can help produce better summaries.

\begin{table}[t]
  \small
  \renewcommand{\arraystretch}{1.2}
\centering
\begin{tabular}{@{}lccc@{}} \toprule
                                  & \textsc{R}-2 & \textsc{\textsc{R}}-3 & \textsc{R}-SU4 \\ \midrule
TAC -- QR-NP (no facet)                  & 13.5   &  5.3    & 19.3    \\
TAC -- QR-NP (faceted)                            & 15.8    & 6.9    & 20.4    \\ \midrule
CL-SciSum -- QR-NP (no facet)                  & 19.4   & 17.2    & 22.6    \\
CL-SciSum -- QR-NP (faceted)                     & 30.2   & 23.9   & 25.7 \\ \bottomrule
\end{tabular}
\caption{The effect of discourse facets on the summarization results on the TAC and CL-SciSum dataset based on QR-NP approach by greedy sentence selection strategy on the identified facets. Other approaches show similar positive trends. Metrics are \textsc{Rouge} F-scores.}
\label{tab:facet}
\end{table}

To better analyze the effect of identifying discourse facets on the overall quality of the summary, we compare the \textsc{Rouge} scores of the summary generated by our approach with and without this step. Table \ref{tab:facet} shows the overall summarization results based on our QR-NP approach when we only use contextualized citations compared with when we use faceted contextualized citations. We observe that grouping citation contexts by their corresponding discourse facet has a positive effect on the quality of the summary on both datasets (17\% and 55\% improvements over TAC and CL-SciSum datasets in terms of \textsc{Rouge}-2, respectively). This is because identifying facets and grouping the contextualized citations by facets, results in a summary that captures the content from all sections of the paper. We observe similar trends for other variants of our approaches; for brevity we only show the results for QR-NP as an illustrative analysis on the effect of identifying discourse facets on the quality of the generated summary.

Finally, an example of the generated summaries by our system (QR-NP-greedy) that uses citation contexts and discourse facets is illustrated in Figure \ref{fig:example-summaries}. We observe that compared with the human summary, the summary generated by our system can capture the significant points of the paper.


\section{Discussion}

Citations are a significant part of scientific papers and analysis of citation texts can provide valuable information for various scholary applications. Our work provides new approaches for contextualizing citations which is a sub-task for enriching citation texts and thus can benefit various bibliometric enhanced NLP applications such as information extraction, information retrieval, article recommendation, and article summarization.  Our work provides a comprehensive new framework for summarizing scientific papers that helps generating better scientific summaries.

We note that our evaluation was based on the \textsc{Rouge} automatic summarization evaluation framework. Automatic evaluation metrics have their own limitations and cannot fully characterize the effectiveness of the systems. Manual or semi-manual evaluation of summarization (e.g. through Pyramid framework) are alternative evaluation approaches that can provide additional insights into the performance of the systems. Yet, due to expense and reproduction issues, most of the standard evaluation benchmarks including TAC and CL-SciSum have been evaluated through \textsc{Rouge}. As it is standard in the field and to be able to compare our results with the related work, we used the \textsc{Rouge} framework for evaluation. We also note that our focus has been on the content quality of the summaries and other criteria such as coherence and linguistic cohesion have not been the focus of our approach. Future work can investigate approaches for improving coherence and linguistic properties of the generated summaries.

\section{Conclusions}

We presented a unified framework for scientific summarization; our framework consists of three main parts: finding the context for the citations in the reference paper, identifying the discourse facet of each citation context, and generating the summary from the faceted citation contexts. We utilized query reformulation methods, word embeddings, and domain knowledge in our methods to capture the terminology variations between the citing and cited authors. We furthermore took advantage of the scientific discourse structure of the articles. We demonstrated the effectiveness of our approach on two scientific summarization benchmarks each in a different domain. We improved over the state-of-the-art by large margins in most of the tasks. While the results are encouraging, the absolute values of some metrics especially in the contextualization task suggest that this problem is worth further exploration. Contextualizing citations is a new task and not only it helps improving scientific summarization, but also it can benefit other bibliometric enhanced end-to-end applications such as keyword extraction, information retrieval, and article recommendation.
%
%


\bibliographystyle{spmpsci}      
\bibliography{ijdl}   


\end{document}